\documentclass[10pt,twocolumn,letterpaper]{article}

\usepackage{iccv}
\usepackage{times}
\usepackage{epsfig}
\usepackage{graphicx}
\usepackage{amsmath}
\usepackage{amssymb}
\usepackage{array}
\usepackage{booktabs}
\usepackage{arydshln}
\usepackage{algpseudocode}
\usepackage{multirow}
\usepackage{multicol}
\usepackage{bm}
\usepackage{caption}
\usepackage{enumitem}
\usepackage[ruled,linesnumbered]{algorithm2e} 

\newcommand{\Rmnum}[1]{\expandafter\@slowromancap\romannumeral #1@}
\newcommand{\tabincell}[2]{\begin{tabular}{@{}#1@{}}#2\end{tabular}}  

\usepackage[breaklinks=true,bookmarks=false]{hyperref}

\iccvfinalcopy 

\def\iccvPaperID{5641} 

\begin{document}

\title{HRGE-Net: Hierarchical Relational Graph Embedding Network for Multi-view 3D Shape Recognition}

\author{Xin Wei\\
	Xi'an Jiaotong university\\
	{\tt\small wxmath@stu.xjtu.edu.cn}
	\and
	Ruixuan Yu\\
	Xi'an Jiaotong University\\
	{\tt\small yuruixuan123@stu.xjtu.edu.cn}
	\and
	Jian Sun\thanks{Corresponding author}\\
	Xi'an Jiaotong University\\
	{\tt\small jiansun@xjtu.edu.cn}
}
\maketitle

\begin{abstract}
	View-based approach that recognizes 3D shape through its projected 2D images achieved state-of-the-art performance for 3D shape recognition. One essential challenge for view-based approach is how to aggregate the multi-view features extracted from 2D images to be a global 3D shape descriptor. In this work, we propose a novel feature aggregation network by fully investigating the relations among views. We construct a relational graph with multi-view images as nodes, and design relational graph embedding by modeling pairwise and neighboring relations among views. By gradually coarsening the graph, we build a hierarchical relational graph embedding network (HRGE-Net) to aggregate the multi-view features to be a global shape descriptor. Extensive experiments show that HRGE-Net achieves state-of-the-art performance for 3D shape classification and retrieval on benchmark datasets. 
	
\end{abstract}

\section{Introduction}
3D shape recognition is an important research area in computer vision. The 3D shapes, including real scanned or CAD objects, retain richer geometric and shape information for recognition than the 2D images captured from a single view. 3D shape recognition plays a critical role in applications such as automatic drive \cite{pylvanainen2010automatic}, archaeology \cite{richards2012kinect}, virtual reality / augmented reality \cite{hagbi2011shape}, etc.

\begin{figure}
	\begin{center}
	\setlength{\belowcaptionskip}{5pt}
		\includegraphics[width=1\linewidth]{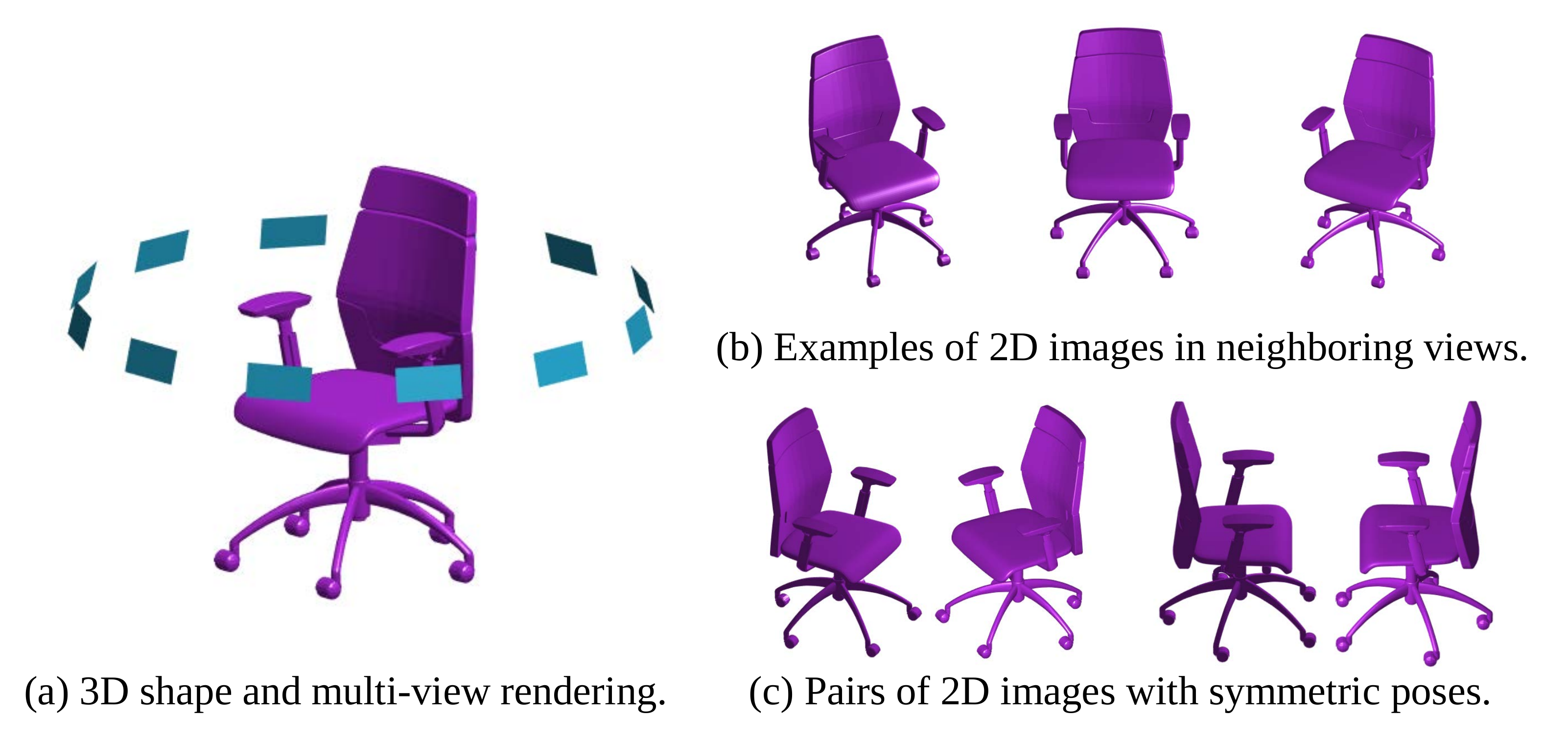}
	\end{center}
	\caption{Illustration of a 3D shape and its multi-view images. The images are rendered from 12 views by camera around the shape as shown in (a). The images from neighboring views are related in both poses and appearance as shown in (b). In (c), the left and right pairs of images are symmetric in poses.}
	\label{fig:figure1}
	\vspace{-0.5cm}
\end{figure}

There have been tremendous advances in research on deep learning-based 3D shape recognition in recent years. According to shape representation, they can be divided into three categories including voxel-based, point-based, and view-based methods. 
\textit{Voxel-based methods} represent a 3D shape by a collection of voxels \cite{mvcnn_multires} in 3D Euclidean space, then build neural networks on voxels to learn the 3D features for recognition \cite{3dshapenet,maturana2015voxnet}. Though they are effective in performance, they commonly have great challenges including the computational complexity, voxel resolution, and data sparsity caused by voxelization of the shape surface. 
\textit{Point-based methods} directly define networks on point clouds or mesh. PointNet \cite{pointnet} is a simple but powerful deep architecture that takes point positions as input. Succeeding methods, e.g., PointNet++ \cite{pointnet2}, SpiderCNN \cite{xu2018spidercnn}, PointCNN \cite{li2018pointcnn}, achieve improved performance for 3D shape recognition. \textit{View-based methods}~\cite{mvcnn,mvcnn_multires,MHBN,rotationnet} recognize shape categories by extracting and aggregating multi-view features, and commonly achieve state-of-the-art performance for shape recognition. However, the challenges are how to project 3D shapes and how to effectively aggregate features learned from multi-view 2D images.

In this work, we focus on the research of view-based 3D shape recognition, and propose a novel network architecture to embed the multi-view features to a global 3D shape descriptor for 3D shape representation. For view-based methods, the simple max-pooling or average-pooling of multi-view features ignores the relations among multi-view images. As shown in Fig.~\ref{fig:figure1}(b), the rendered 2D images captured from neighboring views have strong relation, e.g., the poses and appearance are similar and smoothly changed. Moreover, as shown in Fig.~\ref{fig:figure1}(c), the paired multi-view 2D images in different views are also related, e.g., the examples of paired 2D images are symmetric. Therefore, we believe that the 2D images of neighboring views and pairwise views contain valuable information that could be meaningful for aggregating multi-view features.

Motivated by the above analysis, we propose a novel relational graph network to effectively aggregate the multi-view features for 3D shape recognition. We first construct a relational graph over the multi-view images, and then design a network block of \emph{relational graph embedding} over the graph. This network block explicitly models the pairwise and neighboring view relations among multi-view images respectively by \emph{pairwise relation module} and \emph{neighboring relation module}. Based on this relational modeling, we successively coarsen the relational graph, and design a \emph{hierarchical relational graph embedding network}, dubbed as HRGE-Net, over the graph hierarchy. The learned HRGE-Net can gradually aggregate the multi-view features considering relations among views and produce a discriminative global shape descriptor. 

Compared with traditional view-based methods, this relational graph can effectively explore the relations hidden among the views. 
We evaluate our method on 3D benchmark datasets, and our network achieves state-of-the-art performance, e.g., $95.0\%$ per class and $96.8\%$ per instance classification accuracies on ModelNet40 \cite{3dshapenet}, $77.2\%$ micro-averaged and $63.8\%$ macro-averaged retrieval mAP on ShapeNet Core55 \cite{shapenet}. 

\section{Related Work}

\subsection{View-based 3D Shape Recognition}
Multi-view images of a 3D shape contain rich shape information, and view-based methods commonly lead to higher accuracy for shape recognition compared with voxel and point-based methods. In \cite{gift}, an efficient 3D retrieval system was built by extracting features of multi-view images by CNN and matching two shapes by defining similarity between two sets of view features. In \cite{mvcnn,mvcnn_new}, multi-view image features were pooled across views by max-pooling and then passed through additional network layers to obtain a compact shape descriptor for recognition.

Recently, several works consider the advanced fusion strategy for multi-view feature aggregation. \cite{MHBN} proposed a harmonized bilinear pooling for aggregating the multi-view features from the perspective of patches to patches similarity. In \cite{chen2018veram,han2019seqviews2seqlabels}, the sequential multi-view images were selected and / or aggregated by a RNN with attention for producing a global shape representation. Both of \cite{gvcnn,DSC} investigated the grouping relationship of multi-view features and designed feature pooling on view groups. In \cite{pvrnet}, a point-view network was proposed for integrating the point cloud and multi-view data for joint 3D shape recognition. RotationNet \cite{rotationnet} is one of the state-of-the-art methods for shape recognition, which treated the view index as an optimized latent variable when predicting shape labels. 

These research achieved promising results for 3D shape classification and retrieval. Compared with them, our approach models the multi-view images of a shape as a graph and explicitly learns the relations among the pairwise views and neighboring views by a novel hierarchical relational graph embedding network. As shown in the experiments, it achieves state-of-the-art results for shape recognition.


\subsection{Relational Graph Network}
Modeling the relations among entities or objects is an important task in real-world applications.
Objects constitute the nodes of a graph, and the relations among objects can be modeled as graph edges. Early works \cite{galleguillos2008object,galleguillos2010context,mottaghi2014role} coped with the object relations by a post-processing operation to re-weight the objects. Recently, the relations are modeled in deep learning framework, where LSTM is utilized for sequential reasoning \cite{li2017attentive,stewart2016end}. Relation network~\cite{RN} is a pioneer work for relation reasoning with a simple neural network. It first recognizes objects in the raw input data, and then utilizes a relational reasoning module to reason about the objects and their interactions. Based on~\cite{RN}, \cite{palm2018recurrent} proposed a recurrent relational reasoning module to model the message passing process on graph. In~\cite{hu2018relation}, attention strategy was introduced to relation networks for instance recognition. These works have shown notable performance in applications such as visual QA (question and answer), text-based QA, and dynamic physical systems, justifying the effectiveness of relational modeling of objects / entities. 

Our network is inspired by relation network~\cite{RN,palm2018recurrent}, but with novel designs considering domain knowledge in 3D shape recognition. For example, we take the multi-view features as graph nodes instead of objects, we further design two types of relations, i.e., pairwise and neighboring view relations, and organize these relation models to be a hierarchical deep architecture over gradually coarsened graphs. By ablation study, our relational modules and graph architecture are beneficial for improving the performance of shape recognition.

%

\begin{figure*}
	\setlength{\abovecaptionskip}{-3pt}
	\setlength{\belowcaptionskip}{-5pt}
	\begin{center}
		\includegraphics[width=1\linewidth]{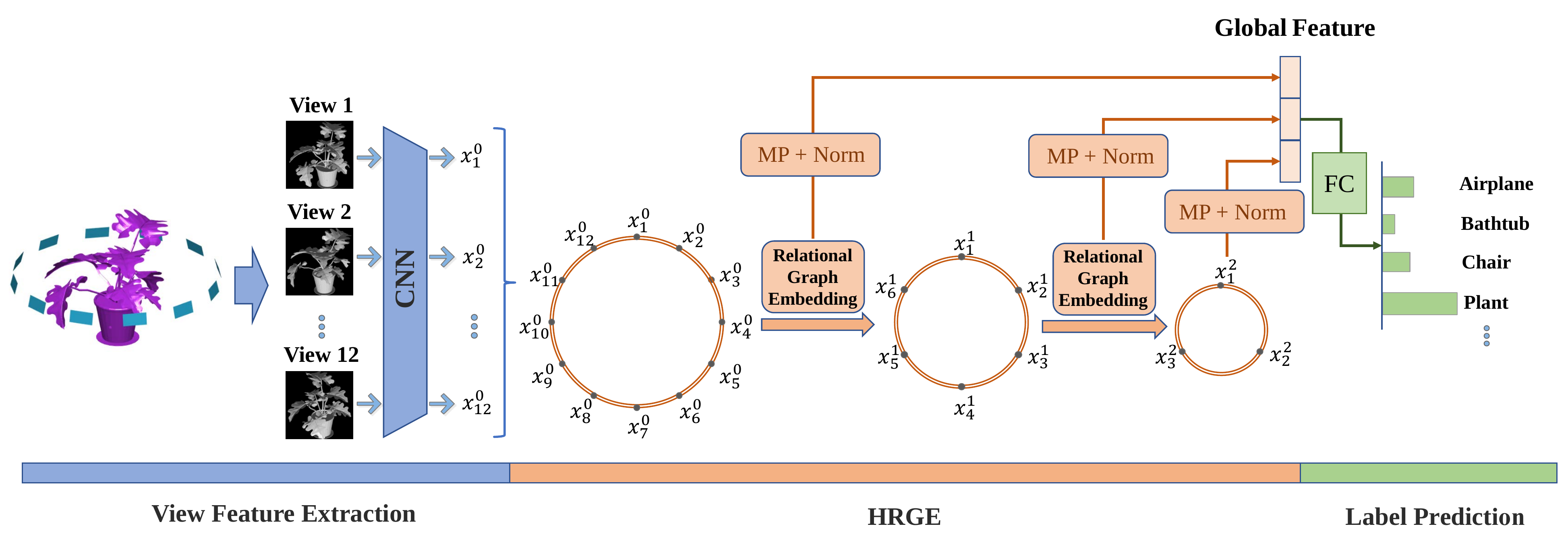}
	\end{center}
	\caption{Pipeline of hierarchical relational graph embedding network. It consists of three stages: the view feature extraction, hierarchical relational graph embedding (HRGE) and label prediction stages. The HRGE stage is proposed to aggregate multi-view features.}
	\label{fig:figure2}
		\vspace{-0.3cm}
\end{figure*}

\section{HRGE-Net}
In this section, we present the details on the design of our HRGE-Net. As illustrated in Fig.~\ref{fig:figure2}, HRGE-Net consists of three stages. In the view feature extraction stage, multi-view features are extracted from projected 2D images of a 3D shape. In the hierarchical relational graph embedding (HRGE) stage, the features are aggregated to be a global feature to represent the shape, which is then sent to the last label prediction stage for predicting its category. 

\subsection{View Feature Extraction}

We project a 3D shape to multi-view 2D images using similar settings as in~\cite{mvcnn,mvcnn_new} by Phong reflection model \cite{phong1975illumination}. Given a 3D object, we first rescale it to fit the fixed viewing volume. Then the virtual cameras are radially symmetric placed, and elevated with 30 degrees around the upright direction. Finally, we render the objects to the virtual camera plane to form a series of images $\{I_i\}_{i=1}^N$ with views indexed by $i$, and the rendered images are with black background. For more details, please refer to \cite{mvcnn,mvcnn_new}.

For the rendered multi-view images $\{I_i \}_{i=1}^N$, we extract their features by a fine-tuned ResNet-50 \cite{he2016deep} network. The network is firstly pre-trained on ImageNet \cite{deng2009imagenet} for image classification, then fine-tuned on the shuffled multi-view images of all the 3D shapes for classification. Finally, the features before the last fully connected layer are taken as multi-view features. All the 2D images from different views share ResNet-50 for feature extraction. For multi-view images $\{I_i \}_{i=1}^N$, after the feature extraction, we derive multi-view features $\{\bm{x}_i \}_{i=1}^N$, which are taken as inputs of our following hierarchical relational graph embedding stage.

\subsection{Hierarchical Relational Graph Embedding}

In this stage, we aim to aggregate the extracted multi-view features $\{\bm{x}_i \}_{i=1}^N$ to be a global 3D shape descriptor. 
As illustrated in Fig.~\ref{fig:figure1}, the rendered images from neighboring views have strong relations (see Fig.~\ref{fig:figure1}(b)) in both poses and appearance. Moreover, we also believe that the rendered images between pairwise views also have strong relations, e.g., the images may be symmetric from some specific paired views (see Fig.~\ref{fig:figure1}(c)). These relations should provide additional discriminative information for shape recognition. Motivated by these observations, we are interested to design a network that can learn to aggregate these multi-view features considering their relations.

\textit{Multi-view Relational Graph.} Given the multi-view features, a graph can be constructed using each view's feature as a node and the view-based relations (will be discussed later) as edges. Since we assume that the virtual cameras lie on a circle around object, therefore the graph is defined on a circle around the object, as illustrated in HRGE stage in Fig.~\ref{fig:figure2}. Our hierarchical relational graph embedding hierarchically aggregates the input $N$ view features ($\{\bm{x}_i \}_{i=1}^N$) from finer graph with $N$ nodes to coarser graphs with decreasing number of nodes (i.e., views) by graph coarsening. Without loss of generality, at level $l$ of the graph hierarchy, we denote the corresponding relational graph as $G^l$ with its graph node feature as $\{\bm{x}_i^l \}_{i=1}^{N_l}$, where $N_l$ is the number of views at level $l$. 

At the level $l$ of graph hierarchy, we next design \textit{relational graph embedding} to aggregate shape features on graph $G^l$ considering the relations of pairwise views and neighboring views.



\subsubsection{Pairwise Relation Module}
This module is responsible for modeling pairwise relation among nodes (i.e., views) in graph $G^l$ to investigate the relations of all pairwise views. 
As shown in Fig.~\ref{fig:figure3}, we set the graph edges as the edges connecting each pair of graph nodes. Then, we define the pairwise relation between nodes $i,j$ of graph $G^l$ as:
\begin{equation}
\bm{r}^l_{ij} = f_{\theta_l}^l([\bm{x}_i^l,\bm{x}_j^l]), i,j = 1,2,...,N_l
\label{eqn:pairwise1}
\end{equation}
where $[\cdot,\cdot]$ denotes the concatenation of two vectors. $f_{\theta_l}^l(\cdot)$ is a relation function with learnable parameter $\theta_l$, aiming at exploring the relation between two nodes of the graph. In our implementation, we design it as a three-layer MLP with 2048 units in each pairwise relation module. 

By Eqn.~(\ref{eqn:pairwise1}), we derive the relations between any two nodes of the graph, then we further gather these relations to achieve the relational feature for each graph node: 
\begin{equation}
R_i^l=\sum_{j\in \Omega(i)}\bm{r}_{ij}^l, i=1,2,...,N_l
\label{eqn:pairwise2}
\end{equation}
where $\Omega(i)$ denotes all nodes that have edges connecting to node $i$. This is similar to the message passing process that collecting information from its connected nodes.

For node $i$, we then design an operation to fuse the relational feature $R_i^l$ with its original feature $\bm{x}_i^l$ as: 
\begin{equation}
\tilde{\bm{x}}_i^l=g_{\phi_l}^l([\bm{x}_i^l,R_i^l]), i=1,2,...,N_l
\label{eqn:pairwise3}
\end{equation}
where $g_{\phi_l}^l(\cdot)$ is a feature fusion function with learnable parameter $\phi_l$. In our implementation, it is designed as a simple fully connected layer with 2048 units. By Eqn.~(\ref{eqn:pairwise3}), node features are updated considering pairwise relations on the graph, such that the updated features are with larger receptive field on the graph. After pairwise relation module, we derive the relational graph $G^l$ with updated node features $\tilde{\bm{x}}_i^l$, which are taken as inputs of the following neighboring relation module.

\subsubsection{Neighboring Relation Module}

We further model the relations among neighboring views, because the neighboring views are related with continuously changed poses and appearance (e.g., Fig. \ref{fig:figure1}(b)).
For the graph $G^l$ with its node feature $\tilde{\bm{x}}_i^l$, we compute the neighboring relation for each graph node $i$ as:
\begin{equation}
\hat{\bm{x}}_i^{l}=
\begin{cases}
h_{\psi_l}^{l}([{\tilde{\bm{x}}_{i-1}}^{l},{\tilde{\bm{x}}_i}^{l},{\tilde{\bm{x}}_{i+1}}^{l}]), & i=2,...,N_l-1\\
h_{\psi_l}^{l}([{\tilde{\bm{x}}_{N_l}}^{l},{\tilde{\bm{x}}_i}^{l},{\tilde{\bm{x}}_{i+1}}^{l}]), & i=1\\
h_{\psi_l}^{l}([{\tilde{\bm{x}}_{i-1}}^{l},{\tilde{\bm{x}}_i}^{l},{\tilde{\bm{x}}_{1}}^{l}]), & i=N_l
\end{cases}
\label{eqn:triple1}
\end{equation}
where $[\cdot,\cdot,\cdot]$ is a concatenation operation, $h_{\psi_l}^{l}(\cdot)$ is a relation function with learnable parameter $\psi_l$, and it is designed to investigate the relations among triplet of neighboring views' features, and maps the high dimensional concatenated features into a low dimension feature space. In our work, we design $h_{\psi_l}^{l}(\cdot)$ as a fully connected layer with 2048 units. 
By Eqn.~(\ref{eqn:triple1}), we fuse the node features with its neighboring node features in the graph, and it can be seen as a convolution operation on the graph with spatial neighborhood of $1 \times 3$.

After updating the node features, we then coarsen the graph $G^l$ by down-sampling the graph nodes with view stride of $s$, achieving a new relational graph $G^{l+1}$ with $N_{l+1} = \frac{N_l}{s}$ nodes, whose nodes features are $\bm{x}_i^{l+1} = \hat{\bm{x}}_{s\times i}^l$, $i = 1, \cdots, N_{l+1}$. We set $s = 2$ in our implementation.

\subsubsection{Summary of Relational Graph Embedding} 
\begin{figure}
	\setlength{\abovecaptionskip}{-3pt}
	\begin{center}
		\includegraphics[width=1\linewidth]{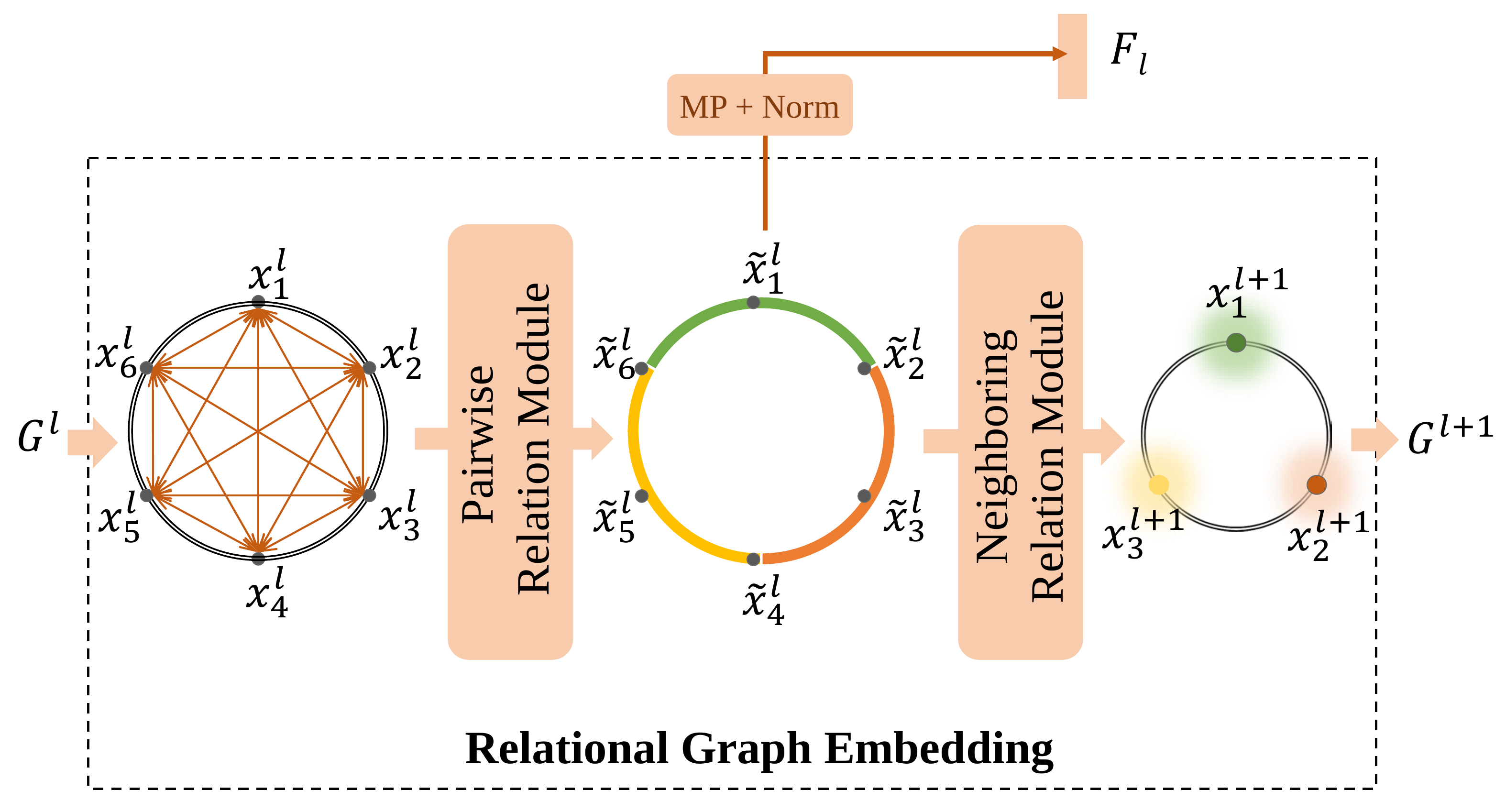}
	\end{center}
	\caption{Relational graph embedding. The node features are first sent to pairwise relation module, followed by neighboring relation module and graph coarsening. The three neighboring views are indicated by green, yellow and red in the middle and right rings. }
	\label{fig:figure3}
		\vspace{-0.3cm}
\end{figure}

In summary, as shown in Fig.~\ref{fig:figure3}, the input multi-view features $\{\bm{x}_i^{l}\}_{i = 1}^{N_l}$ are successively processed by the pairwise relation module and neighboring relation module. The features of each graph node are updated considering features of the other views, and the relations can be automatically learned by training the network. Note that, for simplicity, we implement the relation functions of $f_{\theta_l}^l(\cdot), h_{\psi_l}^l(\cdot)$ and feature fusion function $g_{\phi_l}^l(\cdot)$ with simple neural network layers, and they can also be designed as other complex functions with larger capacity. 

\subsubsection{Hierarchical Relational Graph Embedding}

For one relational graph embedding stage $l$, it embeds multi-view features on a graph $G^l$ to output the updated features on a coarsened graph $G^{l+1}$ with less number of views. We concatenate multiple repetitions of relational graph embedding to be a hierarchical deep architecture, i.e., the HRGE-Net. To retain all the shape features in the hierarchy, in each stage except the last graph $G^L$, we perform max-pooling on multi-view features $\{\tilde{\bm{x}}_i^{l}\}_{i=1}^{N_{l}}$ followed by $l_2$ normalization (Norm) to be a global shape descriptor:
\begin{equation}
F_l = \frac{\hat{F}_l}{||\hat{F}_l||_2}, \hat{F}_l = \mbox{ maxpool}(\{\tilde{\bm{x}}_i^{l}\}_{i=1}^{N_{l}}), l=0,1,...,L-1
\end{equation}
For the last relational graph $G^L$, we directly max-pool and normalize its node features $\{\bm{x}_i^{L}\}_{i=1}^{N_L}$ to get $F_L$, and the final global shape feature is the concatenation of all the global features at all levels: $\bm{F} = [F_0, \cdots, F_L]$.

The proposed hierarchical relational graph embedding is summarized in Algorithm 1 for clarity, where $\theta_l, \phi_l, \psi_l$ ($l = 0, \dots, L-1$) are network parameters to learn. We next present two instances of HRGE-Net with different number of input views. 

\textbf{\textit{6-View HRGE-Net:}} The 6 views of a 3D shape construct a relational graph with 6 nodes. The graph is coarsened once with stride of two, then the HRGE-Net is with relational graph embedding defined over graphs in a hierarchy with 6 nodes and 3 nodes. The final global feature of a 3D shape is concatenation of $F_0, F_1$.

\textbf{\textit{12-View HRGE-Net:}} Taking 12 view features as a relational graph with 12 nodes, the HRGE-Net performs relational graph embedding defined over graph hierarchy with nodes of 12, 6, 3 respectively. Therefore, the final 3D shape descriptor is the concatenation of $F_0, F_1, F_2$. Please refer to Fig.~\ref{fig:figure2} for illustration of 12-view HRGE-Net.

\textbf{\textit{Comparison with CNNs and GNNs:}} Compared with CNNs defined over image grid for recognition, our deep network is also a hierarchical architecture, but models relational pattern of multi-view features defined over a hierarchy of graphs. Compared with conventional graph neural networks (GNN), our graph network is defined over graphs of multi-view features, and models the inter-relations among multi-views motivated by the domain knowledge in multi-view imaging.

\begin{algorithm} 
	\caption{Forward computation of HRGE. $\theta_l, \phi_l, \psi_l$ ($l = 0, \cdots, L-1$) are parameters to learn.} 
	\KwIn{Relational Graph $G^0$, number of the layers $L$} 
	\For {$0 \leq l < L$} { 
		
		\For {each node $i$ of $G^l$}{
			
			\# Learn pairwise relation
			
			$R_i^l\leftarrow\sum_{j\in \Omega(i)} f_{\theta_l}^l([\bm{x}_i^l, \bm{x}_j^l]) $
			
			\# Update view feature
			
			$\tilde{\bm{x}}_i^l\leftarrow g_{\phi_l}^l([\bm{x}_i^l, R_i^l])$	
		}
		
		\# Global shape feature by MP and Norm
		
		$\hat{F}_l\leftarrow maxpool(\{\tilde{\bm{x}}_i^{l}\}_{i=1}^{N_{l}})$
		
		$F_l\leftarrow normalize(\hat{F}_l)$		
		
		\For{each node $i$ of $G^{l}$}{
			
			\# Learn neighboring relation and update view feature
			
			$\bm{x}_i^{l+1}\leftarrow h_{\psi_l}^l([\tilde{\bm{x}}_{(\cdot)}^{l}, \tilde{\bm{x}}_i^l, \tilde{\bm{x}}_{(\cdot)}^l])\text{ by Eqn. (\ref{eqn:triple1})}$ 
		}
		
		\# Graph coarsening by down-sampling
		
		$G^{l+1}\leftarrow G^{l}\downarrow, \{x_i^{l+1}\}_{i=1}^{N_{l+1}} \leftarrow \{x_i^l\}_{i=1}^{N_l}\downarrow $
	}
	\# Global shape feature by MP and Norm
	
	{ $\hat{F}_L\leftarrow maxpool(\{\bm{x}_i^{L}\}_{i=1}^{N_{L}})$\\
		$F_L\leftarrow normalize(\hat{F}_L)$	}		
	
	\# Aggregate the feature in all layers
	
	$\bm{F}=[F_0, ..., F_L]$
	
	\textbf{Output:} global feature $\bm{F}$
	\label{alg:algorithm}
\end{algorithm} 

\subsection{Label Prediction}

In this stage, we take the shape features $\bm{F}$ learned from HRGE as input and predict the label of the 3D shape. In our implementation, the label predictor is taken as a simple fully connected layer followed by softmax operation and a cross-entropy loss function.

\subsection{Network Training}
We train HRGE-Net by two steps similar to \cite{mvcnn}. In the first step, the pre-trained ResNet-50 on ImageNet~\cite{deng2009imagenet} is fine-tuned on all the multi-view images for classification, then the fine-tuned ResNet-50 without its last fully connected layer is taken as view feature extractor. In the second step, we train the whole pipeline including the feature extractor, HRGE, and label predictor for shape recognition by end-to-end training. The gradients of loss w.r.t. parameters of HRGE and ResNet-50 can be computed by auto-differentiation.

When fine tuning the ResNet-50, we use Adam optimizer with weight decay, batch size, epoch number and initial learning rate as $10^{-3}, 64, 30, 5\times 10^{-5}$ respectively. The learning rate is reduced by half every $10$ epochs. When training the whole architecture, we use Adam optimizer with weight decay, batch size, epoch number and initial learning rate as $10^{-3}, 72, 60, 10^{-5}$ respectively. The learning rate is reduced by scale of $0.5$ every $20$ epochs. These training takes about 8 and 16 hours respectively on a NVIDIA GTX 1080 Ti GPU.

\section{Experiments}

In this section, we evaluate our HRGE-Net on benchmark datasets for 3D shape classification and retrieval.

\subsection{Datasets}

\textbf{ModelNet40~\cite{3dshapenet}.} This dataset consists of 12,311 3D shapes from 40 categories, with 9,483 training models and 2,468 test models for shape classification. There are different number of shapes across categories. Various methods have reported their results on this dataset with different shape representations including voxels, point clouds and multi-view images. 

\textbf{ShapeNet Core55~\cite{shapenet}.} This dataset contains a total of 51,162 3D models categorized into 55 classes, which are further divided into 203 sub-categories. The training, validation and test sets consist of 35764, 5133 and 10265 shapes respectively. Different classes have varying number of samples. We select the ``normal'' version of the dataset, i.e., all of the shapes are consistently aligned and normalized to a unit length cube.


\subsection{Evaluation for 3D Shape Classification}

We first evaluate our 12-view HRGE-Net on ModelNet40 for shape classification. We compare with diverse methods. MVCNN \cite{mvcnn} is an effective multi-view shape recognition method based on deep learning, and MVCNN-new \cite{mvcnn_new} improves its results by improving the rendering technique. PVRNet \cite{pvrnet} is an approach fusing the multi-view image and point cloud features. RCPCNN \cite{DSC} exploits the relation among view features using a clustering strategy, and similar strategy is employed in GVCNN \cite{gvcnn}. MHBN \cite{MHBN} aggregates the multi-view features by bilinear pooling. We also compare with models based on points, voxels and mixed representations, including 3DShapeNets~\cite{3dshapenet}, VoxNet~\cite{maturana2015voxnet}, VRN Ensemble~\cite{brock2016generative}, PointNet++~\cite{pointnet2}, Kd-Networks~\cite{klokov2017escape}, MVCNN-MultiRes~\cite{mvcnn_multires}.

The classification results of above methods are presented in Table~\ref{tab:tab1}. It can be observed that we achieve the highest scores for both per class and per instance accuracies. Among the previous methods, MVCNN-new \cite{mvcnn_new} achieved the highest accuracies, but our HERG-Net is superior on both measurements by achieving $1.0\%$ and $1.3\%$ higher scores. Compared with MVCNN-new \cite{mvcnn_new} that aggregates the multi-view features by max-pooling, the improvement demonstrates the effectiveness of our feature aggregation strategy, i.e., fusing the multi-view features gradually with a relational graph. Compared with~\cite{DSC,gvcnn}, our HRGE-Net achieved at least 3\% higher in per instance accuracy.


\begin{figure}
		\setlength{\belowcaptionskip}{5pt}
	\setlength{\abovecaptionskip}{-5pt}
	\begin{center}
		\includegraphics[width=0.8\linewidth]{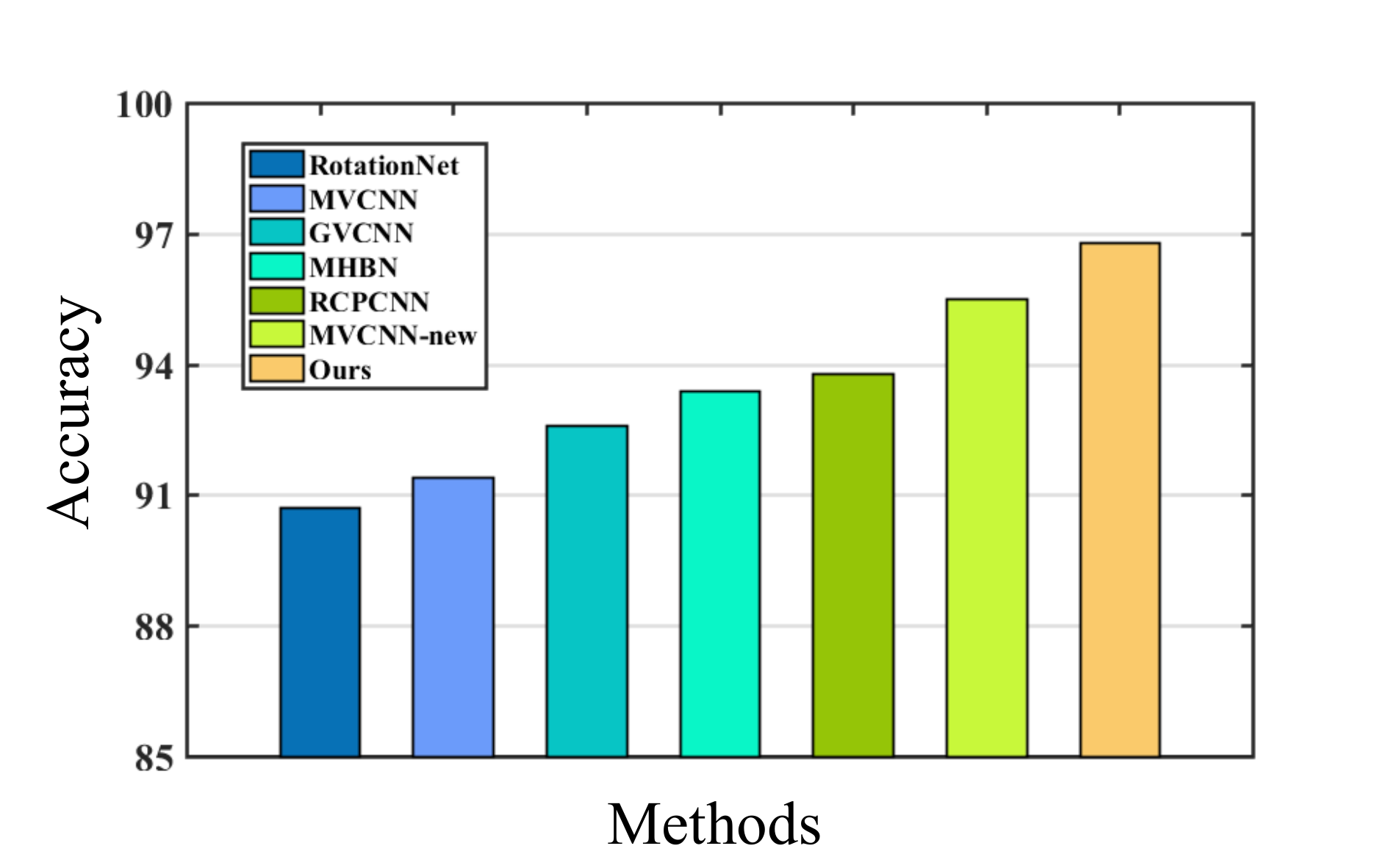}
	\end{center}
	\caption{Per instance classification accuracy on ModelNet40 for some leading view-based methods with 12 views.}
	\label{fig:figure4}
		\vspace{-0.3cm}
\end{figure}

For fair comparison, we also compare with view-based methods with 12 views in Fig.~\ref{fig:figure4}. Compared with the traditional view-pooling methods like MVCNN~\cite{mvcnn}, MVCNN-new~\cite{mvcnn_new}, MHBN~\cite{MHBN}, RCPCNN~\cite{DSC} and GVCNN~\cite{gvcnn}, our HRGE-Net achieves significantly higher accuracy. This improvement shows the effectiveness of our proposed hierarchical relational graph embedding.

RotationNet~\cite{rotationnet} is an interesting approach that estimates poses and tries several variants of camera settings for multi-view projections. Differently, our approach considers how to aggregate the 3D mutli-view features. In Table~\ref{tab:tab1_1}, with ResNet-50 as multi-view feature extractor and 12 views with upright orientation, our HRGE-Net achieves 6.19\% higher in accuracy than RotationNet. By varying the camera poses using 20 views from vertices of a dodecahedron encompassing
the object, the mean accuracy of RotationNet is $94.77\% \pm 1.10\%$, which is also lower than our HRGE-Net. Our HRGE-Net with less number of views which fixed upright orientation achieves almost the same accuracy with RotationNet-Max with best camera poses with 20 views. 

\begin{table}
	\renewcommand\tabcolsep{3.0pt}
	\setlength{\abovecaptionskip}{-5pt} 
	\caption{Shape classification accuracy (in $\%$) on ModelNet40.}
	\begin{center}
		\begin{tabular}{l  c  c c}
			\hline
			\hline\specialrule{0em}{1.2pt}{1.2pt} 
			Method                      &    Input   &   \tabincell{c}{Per Class\\ Acc. }&\tabincell{c}{Per Ins.\\ Acc.}\\\specialrule{0em}{1.2pt}{1.2pt} 
			\hline\specialrule{0em}{1.2pt}{1.2pt} 
			3DShapeNets\cite{3dshapenet} &   \multirow{4}{*}{Voxels} &77.3 &   $-$\\\specialrule{0em}{1pt}{1pt} 
			VoxNet\cite{maturana2015voxnet}                           &        &  83.0  &   $-$\\\specialrule{0em}{1pt}{1pt} 
			VRN Ensemble\cite{brock2016generative}  	  &			&  $-$ &   95.5\\\specialrule{0em}{1pt}{1pt} 
			\hdashline\specialrule{0em}{1pt}{1pt} 
			PointNet$++$ \cite{pointnet2} 	  & \multirow{2}{*}{Points}&$-$&   91.9\\\specialrule{0em}{1pt}{1pt} 
			Kd-Networks \cite{klokov2017escape} 	  &  	      & 88.5    &   91.8\\\specialrule{0em}{1pt}{1pt} 
			\hdashline\specialrule{0em}{1pt}{1pt} 
			MVCNN\cite{mvcnn}         &    \multirow{7}{*}{Images} &90.1    &   90.1\\\specialrule{0em}{1pt}{1pt} 
			MVCNN-new\cite{mvcnn_new}         &          & 94.0   &   95.5\\\specialrule{0em}{1pt}{1pt} 
			MVCNN-MultiRes\cite{mvcnn_multires}              &  	& 91.4  &   93.8\\\specialrule{0em}{1pt}{1pt} 
			PVRNet\cite{pvrnet}         &                          & 91.6   &   93.2\\\specialrule{0em}{1pt}{1pt} 
			GVCNN\cite{gvcnn}      &                               &  90.7  &   93.1\\\specialrule{0em}{1pt}{1pt} 
			RCPCNN\cite{DSC}      &                                &  $-$   &   93.8\\\specialrule{0em}{1pt}{1pt} 
			MHBN  \cite{MHBN}    &                                &   93.1  &   94.7\\\specialrule{0em}{1pt}{1pt} 
			\hdashline \specialrule{0em}{1pt}{1pt}
			HRGE-Net      &    Images                            &   \textbf{95.0}    & \textbf{96.8}\\\specialrule{0em}{1pt}{1pt} 
			\hline\specialrule{0em}{1pt}{1pt}
			\hline\specialrule{0em}{1pt}{1pt}
		\end{tabular}
	\end{center}
	\label{tab:tab1}
		\vspace{-0.3cm}
\end{table}

\begin{table}
	\renewcommand\tabcolsep{10.0pt}
	\setlength{\abovecaptionskip}{-5pt} 
	\caption{Shape classification accuracy (in $\%$) on ModelNet40. All of following models take the ResNet-50 as view feature extractor.}
	\begin{center}
		\begin{tabular}{l  c  c }
			\hline
			\hline\specialrule{0em}{1.2pt}{1.2pt} 
			Method                      &    \# Views   &    Per Ins. Acc.\\\specialrule{0em}{1.2pt}{1.2pt} 
			\hline\specialrule{0em}{1.2pt}{1.2pt} 
			RotationNet-mean    &    20   &   94.77$\pm$1.10\\\specialrule{0em}{1pt}{1pt} 
			RotationNet-max     &    20   &   96.92 \\\specialrule{0em}{1pt}{1pt} 
			\hdashline \specialrule{0em}{1pt}{1pt}
			RotationNet         &    12   &   90.65 \\\specialrule{0em}{1pt}{1pt} 
			HRGE-Net            &    12   &   96.84 \\\specialrule{0em}{1pt}{1pt} 
			\hline\specialrule{0em}{1pt}{1pt}
			\hline\specialrule{0em}{1pt}{1pt}
		\end{tabular}
	\end{center}
	\label{tab:tab1_1}
	\vspace{-0.6cm}
\end{table}

\subsection{Ablation Study}

We next go deeper to justify the effectiveness of each component of our HRGE-Net. We will conduct experiments to show the effectiveness of the pairwise relation module, neighboring relation module, and hierarchical structure. We will also show effects of number of views and feature normalization on performance.

\noindent{\textbf{Effectiveness of pairwise and neighboring relations.}} We first design following baseline architectures. \textit{Baseline}: the global shape feature is the max-pooling of the multi-view image features without graph embedding. \textit{PR}: the multi-view features are firstly sent to the pairwise relation module, and then max-pooled as the global shape features. \textit{NR}: the multi-view features are firstly sent to a neighboring relation module, whose output are then max-pooled as the global shape feature. Note that \textit{Baseline} is in fact the model of MVCNN-new \cite{mvcnn_new}. 
The results of the above architectures are presented in Table~\ref{tab:tab2}. The improved per class and per instance classification accuracies from \textit{Baseline} to \textit{PR} and \textit{NR} are $0.74\%, 0.73\%$ and $0.05\%, 0.65\%$ respectively, which demonstrate the effectiveness of pairwise relation and neighboring relation modules.

\noindent{\textbf{Effectiveness of hierarchical architecture.}} Given 12 input views, we design the following variants of architectures. \textit{HRGE-Net-1L}: our HRGE-Net as shown in Algorithm 1 but with $L = 1$ defined over a hierarchy of two graphs with 12 and 6 nodes. \textit{HRGE-Net}: our full 12-view HRGE-Net. 
As shown in Table~\ref{tab:tab2}, compared with \textit{Baseline}, \textit{HRGE-Net-1L} achieves about $1.02\%$ higher per instance accuracy, showing the effectiveness of our proposed relational graph embedding module. \textit{HRGE-Net} achieves the highest $94.97\%$ per class accuracy and $96.84\%$ per instance accuracy, which are $0.84\%$ and $0.32\%$ higher than \textit{HRGE-Net-1L}.
Further more, to justify the superiority of our neighboring relation module in the hierarchical architecture, we replace the relation function $h_{\psi_l}^l(\cdot)$ in \textit{HRGE-Net} with max-pooling and average-pooling, resulting in \textit{HRGE-Net-MP} and \textit{HRGE-Net-AP}. We also design \textit{HRGE-Net-ID} by designing $h_{\psi_l}^l(\cdot)$ as a mapping for retaining the feature for the node of interest. As in Table~\ref{tab:tab2}, 
\textit{HRGE-Net-MP} is even worse than \textit{HRGE-Net-ID} as a result of destroying the local structure, while \textit{HRGE-Net-AP} performs an average combination of neighboring features and improves the per instance accuracy by $0.12\%$.
Compare with the above two pooling methods, our full \textit{HRGE-Net} can significantly improve the performance of \textit{HRGE-Net-ID} by about $1.07\%$ for per class accuracy and $0.57\%$ for per instance accuracy, which demonstrates the advantages of our proposed neighboring relation module.

\noindent{\textbf{Effectiveness of normalization operation.} In Table~\ref{tab:tab2}, we also present the results of \textit{HRGE-Net (w/o N)} having the same architecture as \textit{HRGE-Net} except that we do not normalize the max-pooled global features when concatenating them. It achieves $0.54\%$ and $0.37\%$ lower scores for per class and per instance accuracies, showing the necessity of feature normalization. It is worth noting that the result of \textit{HRGE-Net (w/o N)} is similar to \textit{HRGE-Net-1L}. This is because the global features at different layers are not in the same scale, and feature normalization is necessary for enabling the network to learn with deeper layers. 

	\noindent{\textbf{Effect of the number of views.}} These above architectures are all based on 12-view HRGE-Net, we further test 6-view HRGE-Net, and it achieves $94.36\%$ per class and $96.39\%$ per instance accuracies, $0.61\%$ and $0.45\%$ lower than the 12-view HRGE-Net. It is notable that the number of views influences the number of layers in our networks.
	
	\begin{table}
		\renewcommand\tabcolsep{7.0pt}
		\setlength{\abovecaptionskip}{-5pt}
		\renewcommand\arraystretch{1.15}
		\caption{Results (in \%) of variants of HRGR-Net for shape classification with different architectures.}
		\begin{center}
			\begin{tabular}{c| l  c  c}
				\hline
				\hline\specialrule{0em}{1.2pt}{1.2pt} 
				\# Views & Method&\tabincell{c}{Per Class\\ Acc. }&\tabincell{c}{Per Ins.\\ Acc.} \\\specialrule{0em}{1.2pt}{1.2pt}
				\hline  \specialrule{0em}{1.2pt}{1.2pt} 
				\multirow{1}{*}{6}  & HRGE-Net &  94.36  &  96.39	               \\\specialrule{0em}{0.8pt}{0.8pt} 
				\hline  \specialrule{0em}{1.2pt}{1.2pt}
				\multirow{9}{*}{12}&Baseline      &   94.00       &   95.50            \\\specialrule{0em}{0.7pt}{0.7pt}  
				&PR            &  94.74       &  96.23           \\\specialrule{0em}{0.7pt}{0.7pt} 
				&NR            &   94.05           &   96.15                \\\specialrule{0em}{0.7pt}{0.7pt}  
				\cdashline{2-4}   \specialrule{0em}{0.9pt}{0.9pt} 
				&HRGE-Net-1L         &      94.13        &    96.52               \\\specialrule{0em}{0.7pt}{0.7pt} 
				\cdashline{2-4}                                         \specialrule{0em}{0.7pt}{0.7pt} 
				&HRGE-Net(w/o N)      &94.43 &96.47                     \\\specialrule{0em}{0.7pt}{0.7pt} 
				\cdashline{2-4}                                         \specialrule{0em}{0.7pt}{0.7pt} 
				&HRGE-Net-MP      &93.60 & 95.99                          \\\specialrule{0em}{0.7pt}{0.7pt} 
				&HRGE-Net-AP    &94.58 & 96.39                         \\\specialrule{0em}{0.7pt}{0.7pt} 
				&HRGE-Net-ID      &93.90 & 96.27                       \\\specialrule{0em}{0.7pt}{0.7pt} 
				\cdashline{2-4}                                        \specialrule{0em}{0.9pt}{0.9pt} 
				&HRGE-Net      &\textbf{94.97}& \textbf{96.84}    \\\specialrule{0em}{0.7pt}{0.7pt} 
				\hline  \specialrule{0em}{1.2pt}{1.2pt} 
				\hline
			\end{tabular}
		\end{center}
		\label{tab:tab2}
			\vspace{-0.7cm}
	\end{table}

	\begin{table*}
		\renewcommand\tabcolsep{6.0pt}
		\setlength{\abovecaptionskip}{-5pt}
		\caption{Shape retrieval results (in $\%$) on ShapeNet Core55 dataset. The first and second top accuracies are presented with bold and underline formats respectively.}
		\begin{center}
			\begin{tabular}{ l  c  c  c  c  c   c  c  c c c}
				\hline
				\hline\specialrule{0em}{1.2pt}{1.2pt}
				\multirow{2}{*}{Method}  &\multicolumn{5}{c }{microALL}   &\multicolumn{5}{c }{macroALL} \\ \specialrule{0em}{1.2pt}{1.2pt}      \cmidrule(r){2-6} \cmidrule(r){7-11}
				&P@N     &R@N      &F1@N    &mAP   &NDCG    &P@R     &R@N      &F1@N
				&mAP  &NDCG\\\specialrule{0em}{1.2pt}{1.2pt}
				\hline  \specialrule{0em}{1.2pt}{1.2pt} 
				ZFDR                       &53.5&25.6&28.2&19.9&33.0&21.9&40.9&19.7&25.5&37.7\\\specialrule{0em}{1pt}{1pt}
				DeepVoxNet     &79.3&21.1&25.3&19.2&27.7&59.8&28.3&25.8&23.2&33.7\\\specialrule{0em}{1pt}{1pt}
				DLAN    & \textbf{81.8} & 68.9& 71.2& 66.3& 76.2&\textbf{61.8}& 53.3& 50.5& 47.7& 56.3\\\specialrule{0em}{1pt}{1pt}
				\hdashline\specialrule{0em}{1pt}{1pt} 
				RotationNet\cite{rotationnet} &\underline{81.0} & 80.1 &\textbf{79.8} &\textbf{77.2}&\textbf{86.5}&\underline{60.2}&63.9&\textbf{59.0}&\underline{58.3}&65.6 \\ \specialrule{0em}{1.2pt}{1pt}
				Improved GIFT\cite{imporved_gift} &78.6 &77.3 &76.7 &72.2 &82.7 &59.2 &65.4 &\underline{58.1}&57.5&\underline{65.7}\\\specialrule{0em}{1pt}{1pt}
				ReVGG   &76.5&\underline{80.3} &77.2  &74.9 &82.8 &51.8 &60.1 &51.9&49.6&55.9\\\specialrule{0em}{1pt}{1pt}
				MVFusionNet       &74.3  &67.7 &69.2  &62.2 &73.2 &52.3 &49.4 &48.4&41.8&50.2\\\specialrule{0em}{1pt}{1pt}
				CM-VGG5-6DB       &41.8  &71.7 &47.9  &54.0 &65.4 &12.2 &\underline{66.7} &16.6&33.9&40.4\\\specialrule{0em}{1pt}{1pt}
				GIFT\cite{gift}   &70.6  &69.5 &68.9  &64.0 &76.5 &44.4 &53.1 &45.4&44.7&54.8\\\specialrule{0em}{1pt}{1pt}
				MVCNN\cite{mvcnn} &77.0  &77.0 &76.4  &73.5 &81.5 &57.1 &62.5 &57.5&56.6&64.0\\\specialrule{0em}{1pt}{1pt}
				\hline\specialrule{0em}{1.2pt}{1.2pt} 
				Ours  & 76.8&\textbf{81.5}&\underline{78.2}&\textbf{77.2}&\underline{85.4}&52.2&  \textbf{71.8} &57.2 & \textbf{63.8}& \textbf{69.6}  \\\specialrule{0em}{1pt}{1pt}
				\hline           
				\hline
			\end{tabular}
		\end{center}
		\label{tab:tab3}
			\vspace{-0.6cm}
	\end{table*}
	
	
	\subsection{Evaluation for 3D Shape Retrieval}
	
	We now evaluate our approach for 3D shape retrieval on ShapeNet Core55 \cite{shapenet}, which is a challenging dataset containing 55 categories and 203 sub-categories. For given 3D shapes, after rendering them to 12 views, we train HRGE-Net for shape classification based on these multi-view images. Then these learned shape features when training classifiers can be taken as the features for retrieval. Following~\cite{rotationnet}, we train two classification HRGE-Nets to respectively predict the shape categories (HRGE-Net-c55) and all the sub-categories (HRGE-Net-c203). For a test data, we send it to HRGE-Net-c55 and take the output before the last fully-connected layer followed by $l_2$ normalization as the 3D shape descriptor for shape retrieval. We first retrieve the shapes using $l_2$-distance of shape descriptors, and drop out these shapes with $l_2$-distance larger than a certain threshold. For the list of retrieved shapes, we further apply HRGE-Net-c203 to predict their sub-category labels and re-rank the list such that shapes in same sub-category to the query shape are ranked higher than others.

	\begin{figure}
		\setlength{\belowcaptionskip}{5pt}
		\setlength{\abovecaptionskip}{-5pt}
		\begin{center}
			\includegraphics[width=1\linewidth]{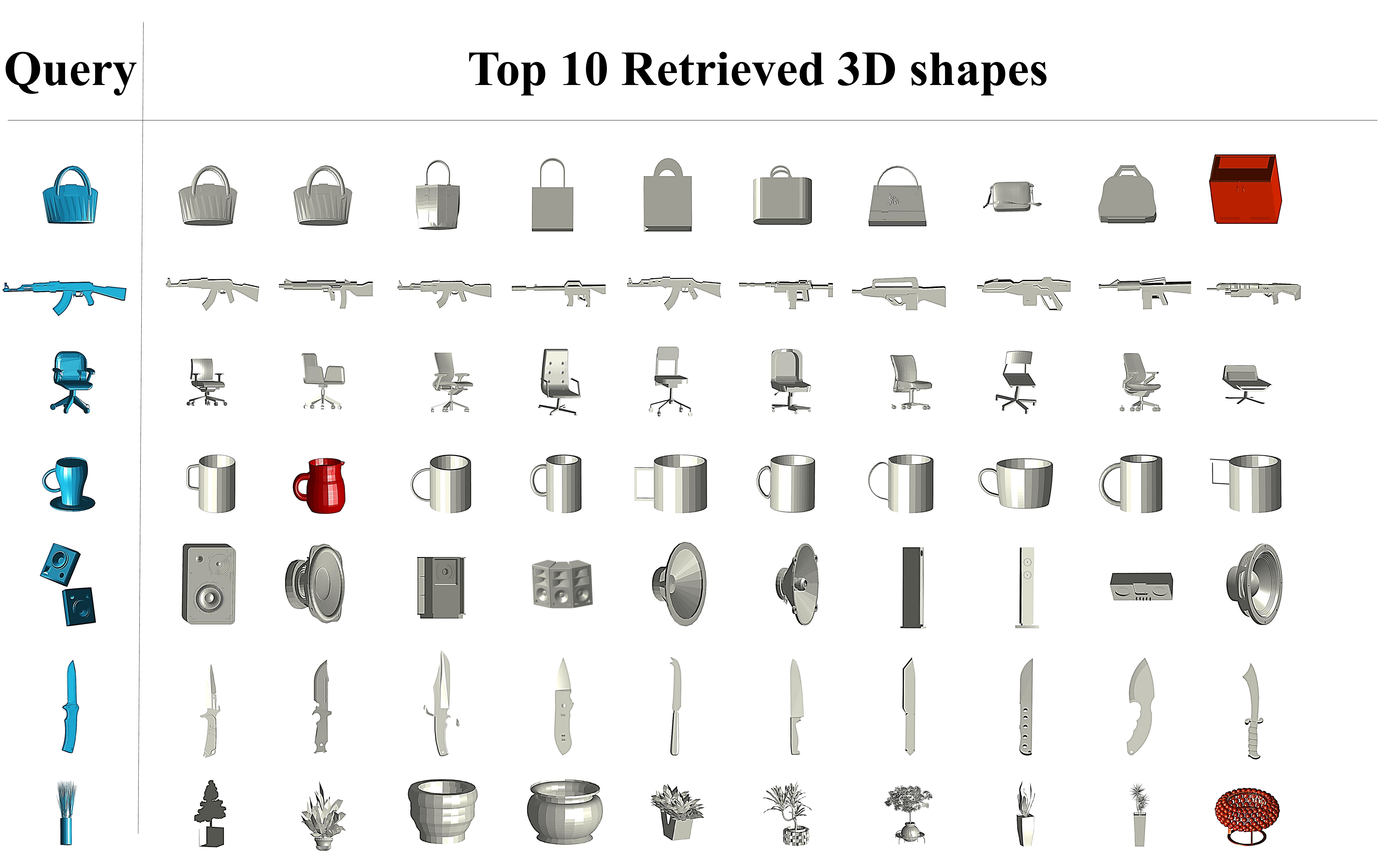}
		\end{center}
		\caption{Retrieval results on ShapeNet Core55 test set. Top 10 matched shapes are shown. Red color indicates failure cases.}
		\label{fig:figure5}
		\vspace{-0.6cm}
	\end{figure}

	We compare with the state-of-the-art methods that attended the track of SHREC'17 for Large-Scale 3D Shape Retrieval~\cite{shapenet} on ShapeNet Core55 dataset. The GIFT~\cite{gift} and Improved-GIFT are multi-view retrieval methods using GIFT and improved GIFT techniques. MVFusionNet takes multi-view images as input and employs a Compact Multi-View Descriptor (CMVD)~\cite{daras2009compact} to generate hand-crafted features which are fused with features from CNN. The method of ReVGG extracts multi-view features by a reduced VGG-M network, and defines similarity with modified Neighbor Set Similarity. CM-VGG55-6DB combines multi-view features to be a global feature and use Clocking Matching algorithm~\cite{lian2013cm} to compute shape dissimilarity. 
	DLAN is a point-set based model by deep aggregation of local 3D geometric features with two well-designed network blocks. ZFDR integrates both visual and geometric information as shape features. DeepVoxelNet designs network on binary voxel grids and the features extracted from the intermediate network layer are taken for shape retrieval. 
	
	As shown in Table \ref{tab:tab3}, our net achieves the highest accuracies for micro-averaged R@N, mAP and macro-averaged R@N, mAP and NDCG@N, and second best for the micro-averaged F1@N and NDCG. Compared with Improve GIFT and MVCNN, HRGE-Net improves mAP and NDCG@N by more than 3.7\% and 2.7\% for microALL, 6.3\% and 3.9\% for macroALL. Furthermore, we outperform the current state-of-the-art method, RotationNet, by 5.5\% in mAP and 4.0\% in NDCG@N for macroALL. This is notable because RotationNet achieves high results by investigating different settings of camera views, while our HRGE-Net relies on a baseline 12-view setting popularly used in 3D shape recognition. We also present examples of retrieval results in Fig.~\ref{fig:figure5}. HRGE-Net can retrieve objects with diverse shapes. For example, in the fifth row that  takes a loudspeaker as query object, we can successfully retrieve various loudspeakers. The red shapes indicate the failure cases, e.g., in the last row that takes potted plant as a query, we retrieve one wrong shape of lamp.

	\section{Conclusion}
	
	In this work, we propose a novel deep relational graph network to aggregate multi-view features for 3D shape recognition. Our proposed network is defined over multi-view relational graphs with a hierarchical architecture, and can be learned to gradually aggregate multi-view features considering pairwise and neighboring relations among views. We have extensively compared our network with previous methods for shape recognition, and showed state-of-the-art performance on benchmark datasets. 
	
	In the future work, we are interested to extend our current network to other camera view settings, e.g., cameras placed on a sphere or dodecahedron around object. It can be achieved by extending our relation modules to be defined on the corresponding graph. We are also interested to introduce attention mechanism into our relational modules to explore better way for feature aggregation. 
	
	{\small
		\bibliographystyle{ieee}
		\bibliography{egbib}
	}


\end{document}